\documentclass[sigconf]{acmart}

\usepackage{booktabs} 

\usepackage[ruled,vlined]{algorithm2e}

\setcopyright{rightsretained}

\acmDOI{}

\acmISBN{}

\acmConference[Preprint]{Preprint}
\acmYear{2021}
\copyrightyear{2021}

\begin{document}
\title[A coevolutionary approach to deep multi-agent reinforcement learning]{A coevolutionary approach to deep multi-agent \\ reinforcement learning}




\author{Daan Klijn}
\affiliation{%
  \institution{Vrije Universiteit Amsterdam}
}
\email{d.s.klijn@student.vu.nl}

\author{A.E. Eiben}
\affiliation{%
\institution{Vrije Universiteit Amsterdam}
}
\email{a.e.eiben@vu.nl}


\begin{abstract}

Traditionally, Deep Artificial Neural Networks (DNN's) are trained through gradient descent. Recent research shows that Deep Neuroevolution (DNE) is also capable of evolving multi-million-parameter DNN's, which proved to be particularly useful in the field of Reinforcement Learning (RL). This is mainly due to its excellent scalability and simplicity compared to the traditional MDP-based RL methods. So far, DNE has only been applied to complex single-agent problems. As evolutionary methods are a natural choice for multi-agent problems, the question arises whether DNE can also be applied in a complex multi-agent setting. In this paper, we describe and validate a new approach based on Coevolution. To validate our approach, we benchmark two Deep Coevolutionary Algorithms on a range of multi-agent Atari games and compare our results against the results of Ape-X DQN. Our results show that these Deep Coevolutionary algorithms (1) can be successfully trained to play various games, (2) outperform Ape-X DQN in some of them, and therefore (3) show that Coevolution can be a viable approach to solving complex multi-agent decision-making problems. 

\end{abstract}

%
%



\maketitle

\section{Introduction}


Recent developments in Reinforcement Learning (RL) made it possible to train intelligent agents that are able to make decisions in complex and uncertain environments. RL methods are mostly MDP-based and often use Neural Networks that serve as function approximators. Using RL, AI's can now master the game of Go \cite{silver2016mastering}, learn to drive a car autonomously \cite{pan2017virtual} and solve a Rubik's cube using a robot hand \cite{akkaya2019solving}. These developments show the importance and capabilities of this class of algorithms. 

\begin{figure}
    \centering
    \includegraphics[width=\linewidth]{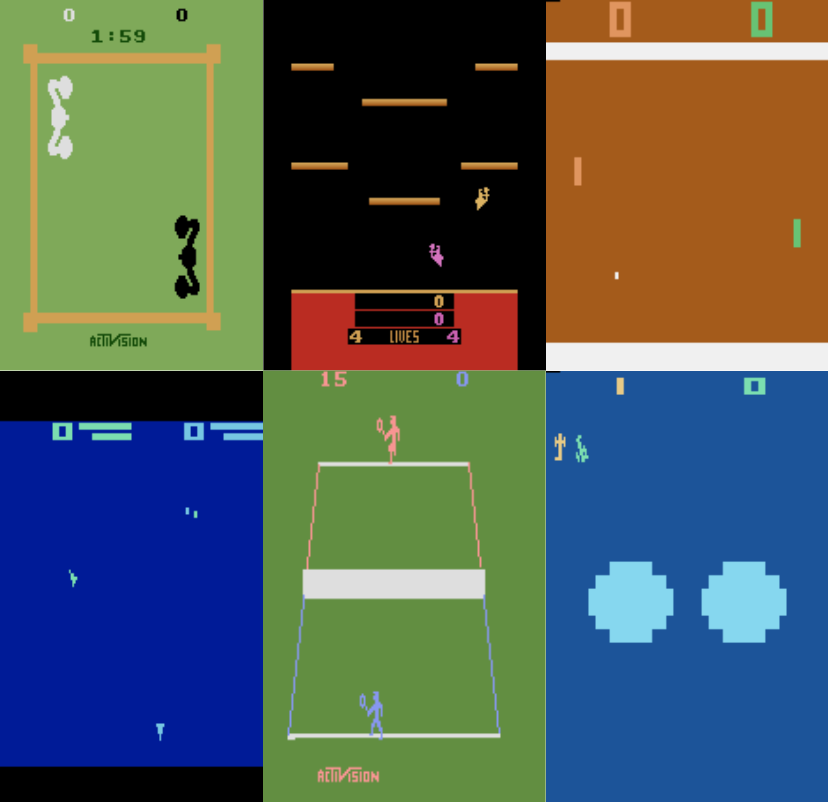}
    \caption{In this work we show that a combination of Deep Neuroevolution and Coevolution is perfectly capable of evolving agents for complex multi-agent problems. The figure above shows a subset of the Atari 2600 games we have used for evaluation and benchmarking of the proposed Coevolutionary algorithms.  }
    \label{fig:my_label}
\end{figure}


The field of Evolutionary Computing also provides a broad range of algorithms that can solve similar sequential decision-making problems. While most RL methods rely on the training of Neural Networks via Stochastic Gradient Descent (SGD), EC offers the possibility to evolve Neural Networks without the usage of a gradient. This gradient-free form of evolving Neural Networks is often called Neuroevolution. Although Neuroevolution has always been a viable solution for sequential decision making (SDM) \cite{igel2003neuroevolution, stanley2002efficient}, only recently has it been proved that Neuroevolution can evolve DNN agents for complex SDM problems as well. In particular, Evolution Strategies \cite{salimans2017evolution} (ES) and Genetic Algorithms \cite{such2017deep} (GA) were used to evolve multi-million-parameter DNN's that were able to achieve competitive results on a range of RL benchmarks while scaling exceptionally well. These findings showed that Neuroevolution is also a viable solution for complex SDM and raises questions about its further potential.


While most RL problems consist of an agent solely interacting with an environment, a subfield called Multi-Agent Reinforcement Learning (MARL) encompasses the problems where multiple agents are present. The goal of MARL is to develop agents that can successfully cooperate or compete with other agents. Similarly, Evolutionary Computation has a class of algorithms that is called Coevolutionary Algorithms. In the same fashion, these algorithms try to evolve individuals who can cooperate or compete with other individuals. Compared to the single-agent flavor of these problems, multi-agent problems tend to be more challenging as they include another uncertainty, namely,  the other agents. 
Due to this, the measured performance of an agent is subjective, as it depends on the other agents' performance. While MARL has been proven to be an effective solution for complex Multi-Agent SDM, Coevolution has never been scaled to high complexity problems.



This paper aims to investigate whether the fields of Coevolution and MARL can be joined to solve complex problems. We do this by proposing a combination of Deep Neuroevolution with a coevolutionary system that
can be employed to solve multi-agent decision-making problems in complex environments. Thus, our main research question is: Is DNE able to scale to complex multi-agent problems? To this end, we develop two Deep NeuroCoevolution algorithms, one based on ES and one using a GA approach inspired by \citet{such2017deep} and \citet{salimans2017evolution}, and test these on eight different games from the PettingZoo benchmark environments \cite{terry2020pettingzoo}. To assess our approach's viability, we compare the evolved agents to the ones delivered by Ape-X DQN. 

\section{Background}

\subsection{Reinforcement Learning}
In brief, the field of Reinforcement Learning (RL) aims to develop intelligent agents that can make decisions in complex and uncertain environments \cite{sutton1998introduction}. These agents should learn to reach a predefined goal by themselves, without any supervision.  This goal is often defined in terms of a numerical reward. RL agents try to learn what behavior is associated with these rewards and finally utilize the learned behavior to reach the predefined goal. 

Most traditional RL methods are MDP-based and borrowed from the field of Dynamic Programming. These methods heavily rely on the number of states in a problem and therefore suffer from the so-called "Curse of dimensionality". Due to this curse, these methods were previously not preferred for complex problems. Recently, this changed as it was discovered that these methods could be combined with ANN's. This lead to solving complex problems that were previously deemed impossible to solve. This combination of RL and ANN's, also called Deep Reinforcement Learning (DRL), initially lead to the development of off-policy methods like the DQN and on-policy methods like the A3C\cite{mnih2016asynchronous}, TRPO\cite{schulman2015trust} and PPO\cite{schulman2017proximal}. While all of these methods rely on the SGD of the underlying ANN's, recent research has shown that Evolutionary Algorithms are also capable of evolving deep ANN's for complex RL problems. 

\subsection{Multi-agent Reinforcement Learning}
So far, we have only discussed situations where a single agent interacts with the environment. Naturally, there are also problems where multiple agents are situated in the same environment and are required to either compete or cooperate with each other. This interaction of agents introduces a range of issues. The biggest issue being, that an agent's performance is always subjective. This is because the performance of an agent partially depends on the behavior of the other agents. Developing agents for these multi-agent SDM problems is often called multi-agent learning.

There is a special class of RL algorithms that deals with these multi-agent problems called Multi-agent Reinforcement Learning (MARL). Often MARL algorithms consist of an altered RL algorithm that learns by playing against itself, older versions of itself, or other concurrently learning agents. Examples of these algorithms are MADDPG \cite{lowe2017multi} (extension of DDPG) and QMix\cite{rashid2018qmix} (Q-learning-based).  Self-play proved to be very effective and allowed RL agents to achieve super-human performance in board games \cite{silver2016mastering}, poker\cite{brown2019superhuman} and many modern video games \cite{vinyals2019grandmaster,berner2019dota}.


\subsection{Evolutionary Algorithms}

Evolutionary algorithms is a class of algorithms that is based on the Darwinian principles of natural selection \cite{eiben2002evolutionary}. These algorithms rely on the same principle: a population of individuals that evolves and undergoes forms of natural selection, leading to a population with a higher fitness. The evolution of the population is generally done by combining two individuals (crossover) or slightly altering a single individual (mutation). From a mathematical perspective, this population of individuals can be seen as a collection of solutions to a fitness function $F$. The evolution of the population will lead to finding solutions that maximize $F$.

In a similar fashion to the traditional RL methods, EA's by themselves are not preferred for certain complex problems, like problems with a large number of inputs. The combination of ANN's and EA's is more capable of solving these types of problems. This combination is often called Neuroevolution. While traditional ANN's are trained using SGD, Neuroevolution utilizes EA's to evolve the weights of an ANN instead. Recently, Neuroevolution showed to be capable of tackling very challenging SDM problems \cite{salimans2017evolution, such2017deep}. Due to this, its excellent scalability and simplicity, Neuroevolution based SDM has gained much popularity in the last few years. In section 3, we describe the mathematical details of these methods. 

\subsection{Coevolution}
EA's use a fitness function to measure the fitness of an individual, this works because the expected return of the function $F$ describes the true fitness of an individual. For multi-agent learning, this statement does not always hold, as for these problems, this function also depends on the other individuals in the environment. For a specific individual, the function $F$ might return a positive number while playing against one individual, while the same function returns a negative number when it plays against another individual. Due to this, it is hard to determine the true fitness of an individual. Coevolution tries to solve this problem. Coevolutionary Algorithms (CoEA's) often do this by using so-called evaluators. Simple CoEA's \cite{popovici2012coevolutionary} first select several individuals  from the population that function as evaluators. Subsequently, these evaluators are used to evaluate each individual in the population. Whenever the range of evaluators is diverse enough, this evaluation method should theoretically give a good approximation of an individual's true fitness. 

Just like EA's, CoEA's also can be extended with a broad range of features. One that proved to be very effective is the Hall of Fame (HoF) \cite{rosin1997new, monroy2006coevolution}. The HoF is simply a collection of winner(s) of each generation of the CoEA. During the training of the CoEA individuals are selected from the HoF and used to evaluate individuals of the current population. Due to this approach, we can make sure that winning traits are never lost, ensuring a smoother convergence.

In a similar fashion to the achievements of the MARL algorithms, Coevolution has also proven to be very effective for multi-agent decision making. Over the years it has provided groundbreaking solutions to optimization problems \cite{hillis1990co}, evolved expert checkers \cite{chellapilla2001evolving} and Othello \cite{szubert2009coevolutionary} agents and learned to play video games based on object information \cite{poppleton2002can, monroy2006coevolution}. In contrast to MARL, Coevolution has never been used for complex SDM like game playing based on pixel inputs.

\subsection{Our contributions}

Both RL and DNE have shown that they are viable options when it comes to developing agents for complex SDM problems. Furthermore, their multi-agent counterparts, MARL and Coevolution, are also good options when multiple agents interact in an environment. While MARL has proven to perform equally well when the problem's complexity increases, it was still unknown whether Coevolution could do the same. In this paper, we show that the combination of Neuroevolution and Coevolution can be scaled to complex multi-agent problems, opening up a new door in the field of deep MARL.

\section{related work}

\subsection{Evolution Strategies}
Evolution Strategies (ES) is a form of black-box optimization that was inspired by the process of natural evolution \cite{eigen1973ingo, schwefel1977evolutionsstrategien}. ES starts with an initial parent parameter vector $\theta$ and then mutates this parent to create a population of new parameter vectors. The fitness of each of the vectors in the population is then evaluated using the objective function $F$ of the problem at hand. ES then computes the next parent by combining the vectors in the population so that vectors with a high fitness have more influence.

There are many ES-varieties. One of the most common varieties is Natural Evolution Strategies (NES) \cite{wierstra2008natural, sehnke2010parameter}. One of the characteristics of NES is that the population is drawn from a moving probability distribution $p_{\psi}(\theta)$ that relies on the parameter $\psi$. NES tries to optimize the parameter $\psi$, such that the expected average return of the objective function $F$ is optimized. In other words, NES tries to optimize $\mathbb{E}_{\theta \sim p_{\psi}} F(\theta)$. This is generally done by performing gradient steps on parameter $\psi$ using an approach that is similar to the one REINFORCE \cite{williams1992simple} takes:
$$
\nabla_{\psi} \mathbb{E}_{\theta \sim p_{\psi}} F(\theta)=\mathbb{E}_{\theta \sim p_{\psi}}\left\{F(\theta) \nabla_{\psi} \log p_{\psi}(\theta)\right\}
$$
The in previous sections described advances \cite{salimans2017evolution} in the field of Neuroevolution rely on a NES variant that is capable of evolving DNN's with milions of paramters. This variant of NES is the same that we will base our research on. This NES uses an isotropic multivariate Gaussian with fixed covariance $\sigma^2$ to create the distribution $p_{\psi}(\theta)$. The default NES setup is reformulated so that the gradient steps can be directly performed with respect to the parameter vector $\theta$:
$$
\nabla_{\theta} \mathbb{E}_{\epsilon \sim N(0, I)} F(\theta+\sigma \epsilon)=\frac{1}{\sigma} \mathbb{E}_{\epsilon \sim N(0, I)}\{F(\theta+\sigma \epsilon) \epsilon\}
$$
The authors approximate the update function above using sampling. The whole setup can be defined by three steps, (1) computing the population using the perturbations $epsilon$, (2) evaluating each individual $\theta + \epsilon$ and (3) updating the current $\theta$. Normalizing the values that are returned by $F$ per generation is also recommended, as that will result in a more stable convergence. Below, the algorithm for this approach is given.

\vspace{0.5cm}
\begin{algorithm}
\textbf{Input:} Learning rate $\alpha$, noise standard deviation $\sigma$, inititial policy parameters $\theta_0$
 
 \For{$t = 0, 1, 2, ...$}{
  Sample $\epsilon_{1}, \ldots \epsilon_{n} \sim \mathcal{N}(0, I)$ \\
  Compute returns $F_{i}=F\left(\theta_{t}+\sigma \epsilon_{i}\right)$ for $i=1, \ldots, n$ \\
  Set $\theta_{t+1} \leftarrow \theta_{t}+\alpha \frac{1}{n \sigma} \sum_{i=1}^{n} F_{i} \epsilon_{i}$
 }
 \caption{Evolution Strategies from \citet{salimans2017evolution}}
\end{algorithm}

\subsection{Genetic Algorithms}
The Genetic Algorithm \cite{holland1992genetic, eiben2003introduction} (GA) is the most common Evolutionary Algorithm. Its most traditional form uses the following workflow; it starts with a population of $n$ individuals and then determines the fitness of each of the individuals. Next, parent selection takes place to generate an intermediate population of $n$ individuals. Here the selection probability of each of the individuals is proportional to their fitness. Finally, crossover and mutation are applied to this intermediate population, generating the next population of again size $n$. Clearly, this GA is not necessarily optimal and can be extended with many features. 

Just like ES, GA's can also be used for Neuroevolution. While ES does not make use of SGD, it still relies on approximations of the gradient, which are similar to finite difference approximations. Therefore this technique cannot be considered entirely gradient-free. GA's are, on the other hand, entirely gradient-free. Due to this difference, it was surprising to see that \citet{such2017deep} was able to train DNN's using a simple GA and obtain competitive results on a range of RL benchmarks. The simple GA used is similar to the traditional GA described above, except that it does not use the crossover operator. Furthermore, it is extended with truncation selection, which makes sure that the top $T$ individuals become the next generation's parents. Also, elitism is used to make sure that the best individual always makes it to the next generation.

\vspace{0.5cm}
\begin{algorithm}
\textbf{Input:} Population size $n$, truncation size $T$,  noise standard deviation $\sigma$, initial population $\mathcal{P}_0$ of policy parameters.

 \For{$t = 0, 1, 2, ...$}{
  
  Compute returns $F_{i}=F\left(\theta_{i}\right)$ for $\theta_{i} \in \mathcal{P}_t$  \\
  Sort $\mathcal{P}_t$ by $F_{i}$ s.t. $\mathcal{P}^0_t = \underset{\theta_i \in \mathcal{P}}{\operatorname{argmax}}(F)$ \\
  Sample perturbations $\epsilon_{1}, \ldots \epsilon_{n-1} \sim \mathcal{N}(0, I)$ \\
  Sample parents $p_{1}, \ldots p_{n-1} \sim \mathcal{P}^{0..T}_{t}$ \\
  Set $\mathcal{P}_{t+1} \leftarrow \{\mathcal{P}^0_t\} \cup \{p_{1} + \sigma\epsilon_{1}, \ldots p_{n-1} + \sigma\epsilon_{n-1}\}$ \\
 }
 \caption{Genetic Algorithm from \citet{such2017deep}}
\end{algorithm}


\section{Methods}

DNE has proved to be an effective solution for complex single-agent decision-making. To determine whether DNE can be scaled to the more challenging multi-agent counterparts of these problems, we introduce a new approach based on Coevolution. Using this approach, we introduce two Deep Coevolutionary Algorithms. As the ES from \citet{salimans2017evolution} and the GA from \citet{such2017deep} proved to be successful, we will base our CoAE's on them. Below we will introduce a Coevoluationary GA and a Coevolutionary ES and describe the technical details of these algorithms. For simplicity's sake, we only address the problem where two homogeneous agents are present. Nevertheless, with some trivial modifications of the algorithms, these problems can be tackled as well.

\subsection{Coevolutionary Evolution Strategies}
We propose an ES that can evolve agents for multi-agent problems. To convert the original ES presented by \cite{salimans2017evolution} to a CoEA, we simply select $k$ previous parents and let them function as evaluators. Then we evaluate each mutated individual $\theta_{t}+\sigma \epsilon_{i}$ against these $k$ evaluators i.e. $\theta_{t}$ to $\theta_{t-k}$. We do this for a couple of reasons:
\begin{enumerate}
    \item Comparing each individual against a range of individuals ensures that the fitness evaluation is more reliable than using a single parent for evaluation.
    \item As comparison is done against some of the parents from previous generations, this indirectly functions as a HoF and therefore ensures that historical traits are not lost.
    \item The fitnesses can be directly compared across individuals as each mutation is evaluated against the same range of parents. This makes it possible to perform a weight update using these fitnesses.
\end{enumerate}
We take the average of the evaluations of each mutation. Next, we use these averages to compute the weight updates, which is done in a similar way to the weight updates of the original ES. The full details of the algorithms can be found in Algorithm \ref{coes}. To the best of our knowledge, this is the first coevolutionary variant of Evolution Strategies ever proposed. 

\vspace{0.5cm}
\begin{algorithm}
\textbf{Input:} Learning rate $\alpha$, HoF size $k$, noise standard deviation $\sigma$, inititial policy parameters $\theta_0$
 
 \For{$t = 0, 1, 2, ...$}{
  Sample $\epsilon_{1}, \ldots \epsilon_{n} \sim \mathcal{N}(0, I)$ \\
  Compute returns $F_{i}= \frac{1}{k}\sum_{s=0}^{k} F\left(\theta_{t-s}, \theta_t+\sigma \epsilon_{i}\right)$ \\ for $i=1, \ldots, n$ \\
  Set $\theta_{t+1} \leftarrow \theta_{t}+\alpha \frac{1}{n \sigma} \sum_{i=1}^{n} F_{i} \epsilon_{i}$
 }
 \caption{Coevolutionary ES}
 \label{coes}
\end{algorithm}

\subsection{Coevolutionary Genetic Algorithm}

The Coevolutionary GA we introduce was greatly inspired by the GA of \citet{such2017deep}. Just as for the ES, we will stay as close to the original design as possible. Again, we replace the original fitness evaluation with an evaluator-based fitness evaluation. For the GA we will use an HoF that stores the best individuals of previous generations. To evaluate individuals, we select the last $k$ elites from the HoF. This makes sure that the fitness approximation is reliable and that historical traits are not lost. Note that for both the CoES and the CoGA, $k$ should not be too high as that will increase the number of evaluations of the function $F$ per individual, which can result in a major increase in runtime if $F$ is expensive to compute. Algorithm \ref{coga} shows the mathematical details of the CoGA.

\vspace{0.5cm}
\begin{algorithm}
\textbf{Input:} Population size $n$, truncation size $T$, HoF size $k$,  noise standard deviation $\sigma$, initial population $\mathcal{P}_0$ of policy parameters.

 \For{$t = 0, 1, 2, ...$}{
  
  Compute returns $F_{i}= \frac{1}{k}\sum_{s=1}^{k} F\left(\mathcal{P}_{t-s}^0, \theta_i \right)$ \\
  for $\theta_{i} \in \mathcal{P}_t$  \\
  Sort $\mathcal{P}_t$ by $F_{i}$ s.t. $\mathcal{P}^0_t = \underset{\theta_i \in \mathcal{P}}{\operatorname{argmax}}(F)$ \\
  Sample perturbations $\epsilon_{1}, \ldots \epsilon_{n-1} \sim \mathcal{N}(0, I)$ \\
  Sample parents $p_{1}, \ldots p_{n-1} \sim \mathcal{P}^{0..T}_{t}$ \\
  Set $\mathcal{P}_{t+1} \leftarrow \{\mathcal{P}^0_t\} \cup \{p_{1} + \sigma\epsilon_{1}, \ldots p_{n-1} + \sigma\epsilon_{n-1}\}$ \\
 }
 \caption{Coevolutionary Genetic Algorithm }
 \label{coga}
\end{algorithm}

\section{Experiments}

This section aims to answer our main research question: \textit{Is DNE able to scale to complex multi-agent problems?} We do this by training the algorithms that were proposed on a range of benchmark problems. We evaluate our trained agents by letting them play against a random agent i.e. an agent that performs randomly chosen actions. Finally, we compare the performance of our agents versus the random agent to the performance of a common RL algorithm against a random agent and report the outcome. The code that was used for these experiments can be found on GitHub \footnote{\url{https://github.com/daanklijn/neurocoevolution/}}
.

\subsection{Network architecture and hyperparameters}

We test our deep coevolutionary methods by evolving a DNN similar to the large DQN used by \citet{mnih2015human}. We chose this architecture as it is by far the most used architecture for the Atari 2600 benchmarks and because it is also used by the ES from  \citet{salimans2017evolution} and the GA from \citet{such2017deep}. This network has three convolutional layers with respectively  32, 64, and 64 channels, followed by a single hidden layer of 512 nodes. The convolutional layers use filters of sizes 8x8, 4x4, and 3x3 combined with strides of sizes 4, 2, and 1. All of these layers use the Rectified Linear Units (ReLU) as their activation function. We use virtual batch normalization (VBN) \cite{salimans2016improved, salimans2017evolution} to ensure that the evolved agents are diverse enough. Without VBN, many evolved agents were not diverse and always performed the same actions, regardless of the input. It was, therefore, interesting to see that \citet{such2017deep} was able to evolve effective agents without the usage of VBN.

In terms of hyperparameters, we again did not diverge much from the parameters used by \citet{salimans2017evolution} and \cite{such2017deep}. While both of these papers use one billion game frames for training, we found that we could answer our research question using only 200 million game frames in combination with a slightly higher noise standard deviation and a smaller population size. Both papers use a population size of 1000 individuals and a noise standard deviation of 0.002 and 0.02 for the GA and ES, respectively. We were able to get the best results using a population size of 200 and noise standard deviations of 0.005 and 0.05. Note that we decided to use different HoF sizes for both the CoES and CoGA. For the CoGA we evaluate each individual against three individuals from the HoF, while for the ES we only evaluate against a single individual from the HoF. This was done because the CoES is less sensitive to the subjectiveness of the evaluations as the CoES takes the average of the fitnesses of all individuals to determine the weight update.  The complete list of hyperparameters we used can be found in Table \ref{hyperparam_table}.

\begin{table}[H]
    \centering
    \begin{tabular}{@{}p{4cm}|p{2cm}p{2cm}@{}}
    \toprule
        Hyperparameter & Co-ES & Co-GA \\
    \midrule
        Population size $n$ & 200     & 200          \\
        Noise standard deviation $\sigma$  & 0.05     & 0.005\\
        HoF size $k$ & 1              & 3\\
        Learning rate $\alpha$ & 0.1 & - \\
        Truncation size $T$   & -                & 5\\
    \bottomrule
    \end{tabular}
    \caption{Hyperparameters that were used for the Co-GA and Co-ES.}
    \label{hyperparam_table}
\end{table}

\subsection{Atari environments and preprocessing}
Most RL research utilizes the Atari environments from OpenAI Gym \cite{brockman2016openai} to benchmark algorithms. Unfortunately, these benchmarks do not include multi-player games. Recently, another collection of Atari benchmarks called PettingZoo has been introduced for multi-agent reinforcement learning \cite{terry2020pettingzoo}. To evaluate the CoEA's, we test our algorithms on eighth of multi-player Atari games from this collection. 

We apply a range of preprocessing steps to the frames outputted by the environments before being fed into the DNN. The preprocessing that we apply is similar to the one done by \citet{mnih2013playing} and consists of:
\begin{itemize}
    \item Rescaling the frames to a size of 84 by 84 pixels.
    \item Stacking frames in batches of four.
    \item Skipping three out of each four frames. Note that we still report the original number of frames in our results.
    \item Standardizing the values that are retrieved from the environment.
\end{itemize}
Next to that, we also add a form of agent indication \cite{gupta2017cooperative} as it is essential for each agent to know whether it is player one or player two. We implement this by adding two more channels to each input, containing a one-hot encoding of the player's ID. Due to this, the final input shape of the network is 84 by 84 by 6. 

In most of the multi-player games that we use, one player has an advantage over the other player. For instance, in the game Pong, the first player always gets to serve the ball first. This advantage can lead to assuming that one agent is "better," while this might be due to the agent being player 1. To counter this, we let each agent always plays two games against another agent. One game where the first agent plays as player one and the second agent player two, and the second game vice versa. 

\begin{figure*}
    \centering
    \includegraphics[width=\textwidth]{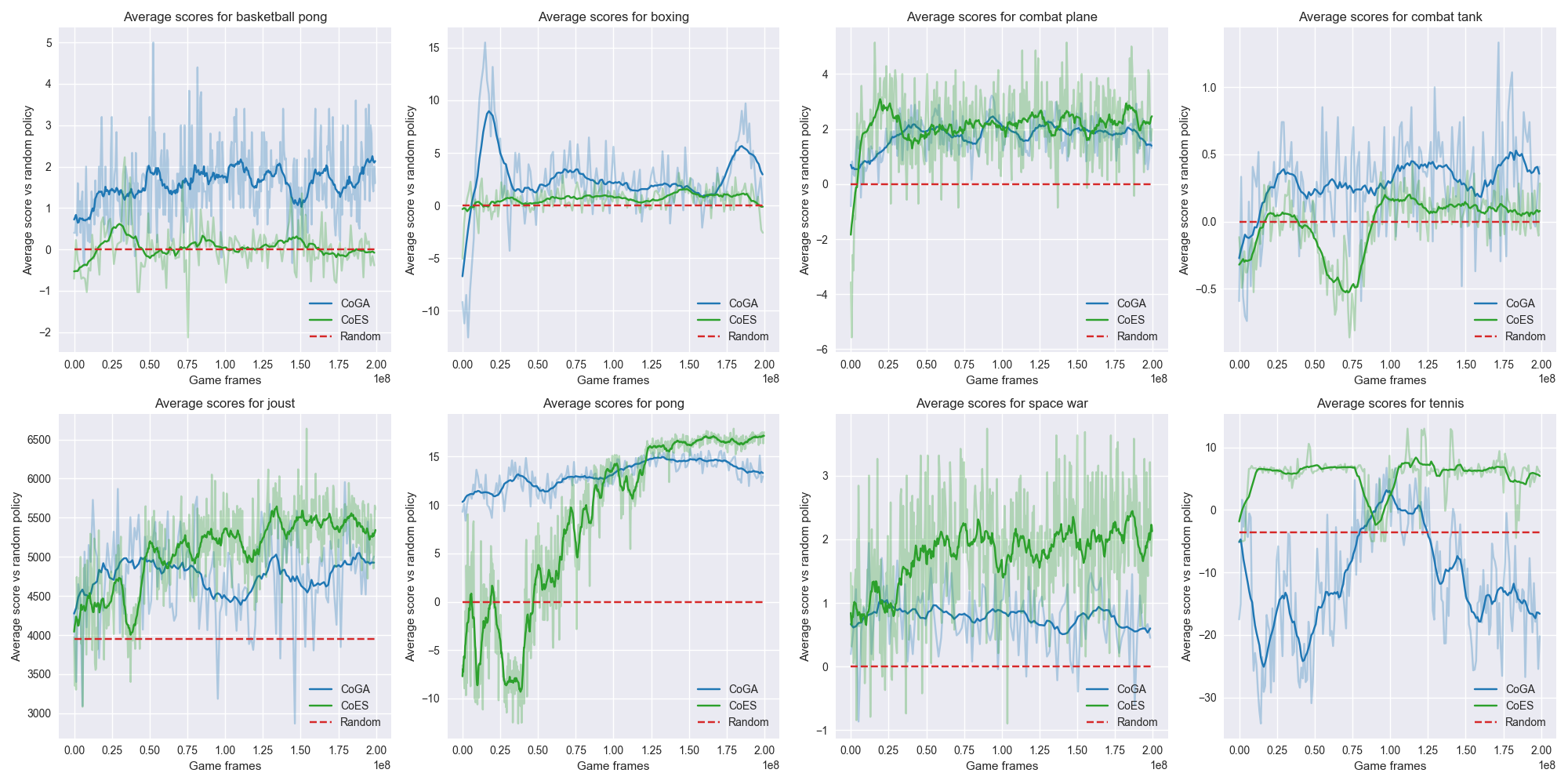}
    \caption{The figures above show the convergence of the average scores of the CoGA and CoES across all generations and games. Each of the data points is based on 50 different games against a random policy. The lines with a low opacity represent the real data points, while the non-transparent lines display a rolling average of these average scores. We have also included a red line indicating the random agent's performance against another random agent. }
    \label{fig:convergence_plots}
\end{figure*}

\subsection{Results}

We trained both the Co-ES and the Co-GA on eight different games, whereas we start learning from scratch for each of the games. We capped the training for each of the games at 200 million game frames. Due to the skipping of frames, this is equal to 50 million training frames. This was equal to around 500 generations for the Co-ES and around 160 generations for the Co-GA. This difference in generations is caused by the Co-GA evaluating each individual against three individuals from the HoF, while the Co-ES only evaluates each individual against a single parent. After each generation, we evaluate the algorithms by playing 50 games against a random agent. For the GA we use the individual with the highest fitness for evaluation. The ES just uses the parameters $\theta_t$ for each generation $t$ for evaluation. The average reward across these 50 games is then used to evaluate the algorithms' performance after each generation. The convergences of both the CoGA and CoES can be found in Figure \ref{fig:convergence_plots}.  To determine each algorithm's final performance, we evaluate each agent again after the 200 million timesteps of training. This time, we let the agent play 250 games against the random agent. The average reward of these 250 games is reported in Table \ref{tab:result_table}. Furthermore, we have included videos of selected gameplays of both the CoES and CoGA \footnote{ \url{https://youtube.com/playlist?list=PLdGO0x90WS8eCCGA1w4jOPLz9ahWSmZEw}}

.

\begin{table}[H]
    \centering
    \begin{tabular}{@{}l||l|lll@{}}
    \toprule
         & Random & Ape-X DQN & Co-ES & Co-GA \\
    \midrule
        Game frames & - & 80M & 200M & 200M \\
        Training frames & - & 20M & 50M & 50M \\
        Generations & - & -  & 500 & 160 \\
        Wall time & - & - & 6h & 6h \\
    \midrule
            Basketball Pong & 0.0 & \textasciitilde 1.0                & 0.0            & \textbf{2.1}     \\

        Boxing          & 0.0  & \textbf{80.0}     & 0.7            & 1.0     \\
        Combat Plane    & 0.0  & -2.0              & \textbf{2.5}  & 1.8     \\
        Combat Tank     & 0.0   &0.3                & 0.1           & \textbf{0.4}     \\
        Joust           & 3951.0 &3846.0            & \textbf{5362.2} & 4910.0     \\  

                Pong            & 0.0 & \textbf{20.5}     & 17.6          & 15.8  \\

                Space war       & 0.0 &1.1                & \textbf{1.8}           & 0.8     \\
        Tennis          & -3.6 & \textbf{22.5}     & 6.9           & -20.2     \\
    \bottomrule
    \end{tabular}
    \caption{Performance of Ape-X DQN, Coevolution ES and Coevolution GA against a Random Policy for various games. The Ape-X DQN results are derived from \cite{terry2020multiplayer} and are based on the last 20 checkpoints for each game.}
    \label{tab:result_table}
\end{table}

Based on the results from Table \ref{tab:result_table} and Figure \ref{fig:convergence_plots} we see that the CoES and CoGA surpass the scores of the Ape-X DQN in 3 and 4 games, respectively. Therefore, we can conclude that for these games the CoEA's are superior. Next to that, we see that both the CoES and CoGA can significantly outperform the random policy in almost all games. Note that we only used 200 million game frames, while the original ES and GA used a billion or more. Due to that and the increasing trends in the convergence plots, we believe that the current benchmark performances are only a mere reflection of the true potential when given enough data. 

The agents' gameplays show that they have successfully evolved a range of skills that lead to being superior in some games. An overview of these skills can be found below.

\begin{itemize}
    \item \textbf{Pong} - The agent learned to track the ball using the paddles. This lead to blocking shots and scoring points.
    \item \textbf{Combat Plane} - The agent learned to chase the other agent. This in combination with shooting lead to scoring multiple points.
    \item \textbf{Joust} - The agent learned to jump on top of the other player, which is essential to score points.
    \item \textbf{Boxing} - The agent learned to dodge the adversary's hits by moving backward, which prevented losing points.
\end{itemize}

While the algorithms' performance is often superior, we also discovered that these algorithms are prone to fall into local optima during the training process. For most games, these local optima consisted of reaching a point where the agent's interactions get stuck in a loop, and no points are scored and last until the end of the game. For Pong this would mean that both agents position themselves exactly at the same height and let the ball bounce between both of them without moving the peddles. For Basketball Pong a similar thing was seen, where one agent would just keep playing and catching the ball so that the other agent could never get the ball and score a point. For Boxing and Combat Tank, both agents would not move at all, so none of them would lose points. For most games, these local optima could be avoided by increasing the noise standard deviation. Even with the increased noise standard deviation, the CoES still could not get out of these local optima for the games \textit{Combat Tank} and \textit{Boxing}. 

It is interesting to see that the EA's perform particularly badly in a niche of environments where the agent's has to take the initiative to score points (Combat tank and Boxing). We believe that this is due to the sparseness of the rewards, when performing random moves. For instance, for Combat Tank, a player needs to perform a reasonably complex combination of actions to obtain a reward. First, it needs to steer towards the direction of the opponent. Then, drive until the opponent in the tanks firing range, steer again, to make sure the tank's canon is pointed in the direction. Finally, it needs to perform the fire action. This, while taking into account that the opponents' moves are non-stationary, makes this a hard policy to evolve. In contrast to that, algorithms that do not treat these problems as a black-box, but instead learn Q-values directly from the game frames, should be more capable of learning these games. This is confirmed by the fact that the Ape-X DQN does perform quite well in Boxing while our EA agents are only slightly better than a random policy.

The training of each of the algorithms took around 6 hours on a machine with 96 CPU cores. There is no reason to believe that the CoES and CoGA are not as scalable as their single-agent counterparts, as the weight updates can still be compressed in the same fashion. Therefore we assume that the same can be achieved in less than one hour when one would distribute the algorithms over 10 of these machines. When it comes to runtime, the coevolutionary algorithms are slower than the original GA and ES. The biggest bottleneck of these algorithms causes a large part of this: the NN forward passes. While the single-agent environments only need to determine the action of a single agent for each frame, the multi-agents need to do this for multiple agents. Therefore in our specific case, we had to perform two NN forward passes for each game frame, instead of a single forward pass. Next to that, we also found that the multi-agent PettingZoo Atari environments are slower than the OpenAI Gym single-agent environments.

\section{Discussion \& Conclusion}
The paper introduces a deep coevolutionary approach that can be applied to challenging multi-agent problems. Following a recent trend, we combine Neuroevolution with a more traditional paradigm from the field of Evolutionary Computing. Coevolution is used together with Neuroevolution and EA's to coevolve agents for complex multi-agent problems. We demonstrated this approach's effectiveness by coevolving effective agents for multi-player Atari games using both ES and a GA. A comparison showed that the results of both Coevolutionary algorithms can be superior to those of Ape-X DQN for several of the games.

Why Coevolution performs so well on these challenging problems compared to the Ape-X DQN is not fully understood. We hypothesize that this might be due to the non-stationarity introduced by the multi-agent aspect of these problems. As MDP's rely on stationarity transition probabilities, MDP-based methods like DQN's might not be the best choice for these non-stationary problems. As EA's do not rely on such assumptions, they might adjust better to the non-stationary nature of multi-agent learning.

Yet, these results are based on a single type of benchmark and are only compared to a single algorithm from the field of RL. We believe that Coevolution can be a great addition to modern multi-agent RL, but also recognize that there are still many more problems and benchmark to expose Coevolution to. Problems of a more cooperative nature or problems containing more than two agents are some of them. 

The simplicity of the algorithms used leaves another research gap. As \citet{such2017deep} already mentioned, the extension of GAs can lead to dramatic performance increases. The same thing can be said about Coevolution. A myriad of features for CoEAs have been developed over the years that can increase the performance of CoEAs. Our CoEAs were of a simple nature and only included a HoF and a form of elitism. As done in \cite{such2017deep}, they could also be used together with Novelty Search, which might reduce the possibility of falling into local optima. Other interesting extensions could be the Pareto Archive \cite{ficici2001pareto} or the usage of multiple populations instead of a single one \cite{popovici2012coevolutionary}.

Our work shows that yet again, an effective paradigm from the field of Evolutionary Computing can be transferred to the field of RL and can be used to solve complex decision-making problems. Coevolution proved to be a viable and sometimes superior approach to solving complex multi-agent decision-making problems. We hope that these findings will open a new door in the field of Deep MARL and lead to the coevolving of many more effective agents.




\bibliographystyle{ACM-Reference-Format}
\bibliography{main} 

\end{document}